\documentclass{article}
\usepackage{ijcai26}

\usepackage{times}
\usepackage{soul}
\usepackage{url}
\usepackage[hidelinks]{hyperref}
\usepackage[utf8]{inputenc}
\usepackage[small]{caption}
\usepackage{graphicx}
\usepackage{amsmath}
\usepackage{amssymb}
\usepackage{rotating}
\usepackage{mathtools}
\usepackage{amsthm}
\usepackage{booktabs}
\usepackage{algorithm}
\usepackage{algorithmic}
\usepackage[switch]{lineno}

\usepackage{multirow}
\usepackage[table]{xcolor}
\usepackage{colortbl}
\usepackage{subcaption}
\usepackage{listings}

\title{Reinforcement Learning for Computer-Use Agents with Autonomous Evaluation\thanks{Accepted to the 4th International Workshop on Generalizing from Limited Resources in the Open World (GLOW @ IJCAI 2026): \url{https://glow-ijcai-2026.github.io/glow-ijcai-2026/}.}}

\author{
Marta Sumyk$^1$
\and
Oleksandr Kosovan$^1$\\
\affiliations
$^1$Ukrainian Catholic University, Lviv, Ukraine\\
\emails
sumyk.pn@ucu.edu.ua, o.kosovan@ucu.edu.ua
}

\begin{document}

\maketitle

\begin{abstract}
Computer-Use Agents (CUAs) execute high-level user goals by perceiving and acting directly within graphical user interfaces. However, reinforcement learning for CUAs remains difficult because open-ended desktop environments rarely provide scalable, machine-readable reward signals: task success is often visually grounded and hard to specify with handcrafted reward functions or dense manual labels.

We propose an RL fine-tuning framework that uses autonomous vision-language evaluation as a scalable supervision signal for GUI agents. Given a final screenshot and the original instruction, a Vision-Language Model judges task completion and provides terminal feedback without task-specific heuristics or manual labels during policy optimization.

Because autonomous evaluators are imperfect, we model their feedback as a noisy binary reward channel and derive a noise-corrected reward estimator for Proximal Policy Optimization. Experiments across macOSWorld, Windows Agent Arena, and OSWorld show that corrected evaluator rewards outperform both zero-shot baselines and raw evaluator rewards, improving success rates by an average of 12.6 percentage points over zero-shot performance and 5.1 points over raw evaluator fine-tuning. These results suggest that autonomous evaluation can serve as a practical reward signal for RL in GUI environments when evaluator noise is explicitly modeled and corrected.
\end{abstract}

\begin{figure}[t]
    \centering
    \includegraphics[width=\columnwidth]{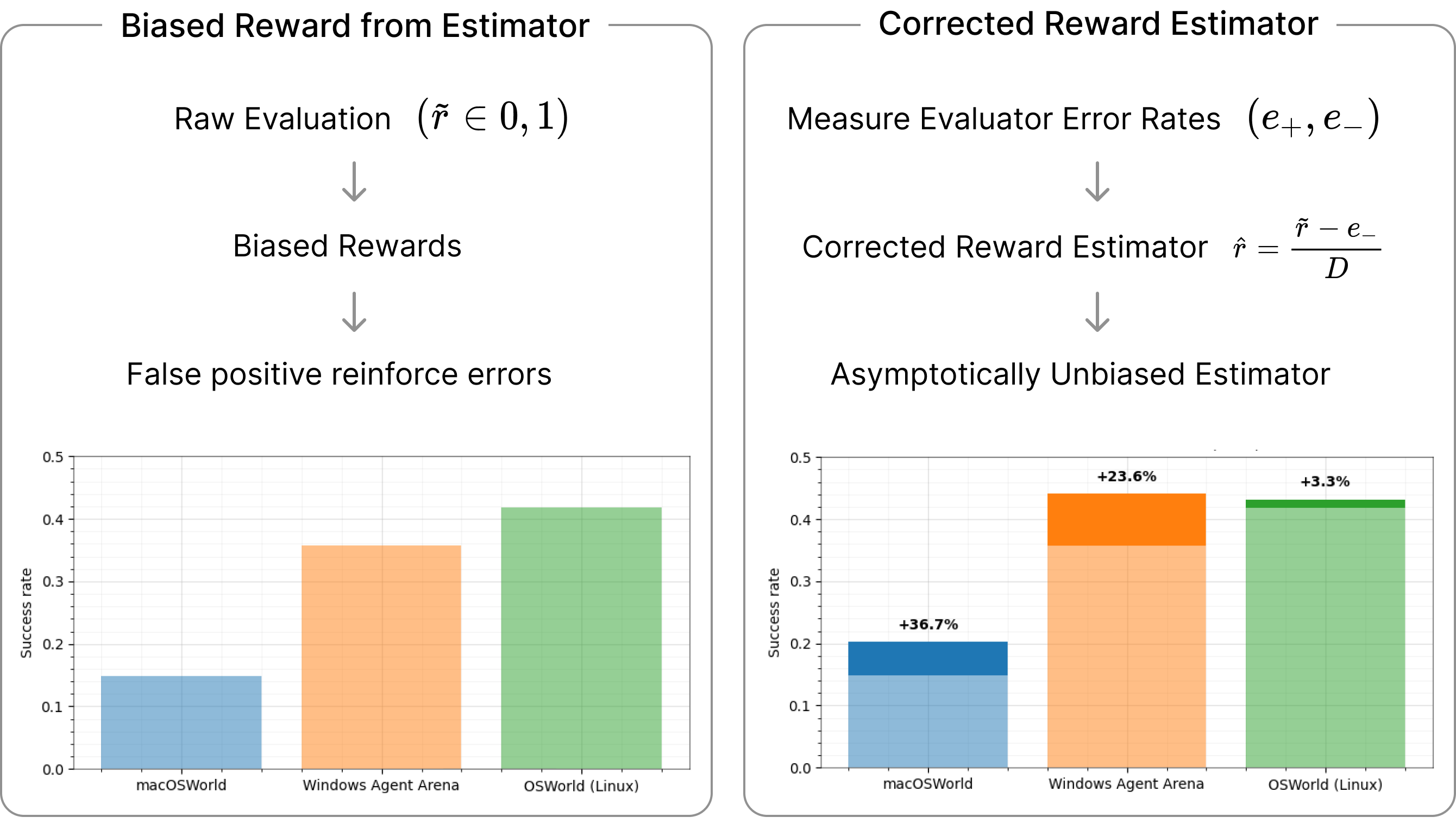}
    \caption{Bias correction framework for autonomous evaluation rewards.}
    \label{fig:reward_correction_schema}
\end{figure}

\section{Introduction}

Computer-Use Agents (CUAs) aim to autonomously operate graphical user interfaces (GUIs) from natural-language instructions using visual observations and executing actions~\cite{liu2025infiguiagentmultimodalgeneralistgui,sun2025seagentselfevolvingcomputeruse}. Recent progress shows promising generalization across applications and operating systems, positioning CUAs as a foundation for service-agnostic desktop automation~\cite{wang2025uitars2technicalreportadvancing,sager2025comprehensivesurveyagentscomputer}. Despite these advances, current CUAs remain unreliable in unconstrained, real-world settings. Specifically, on the OSWorld benchmark~\cite{xie2024osworldbenchmarkingmultimodalagents}, state-of-the-art desktop agents achieve success rates of only around $60\%$, highlighting a significant gap between current capabilities and practical deployment. This gap motivates the need for more effective learning mechanisms to improve robustness and task success in real-world desktop environments.

Reinforcement learning (RL) provides a principled framework for improving agent behavior by enabling exploration, recovery from errors, and learning through interaction rather than reliance on static demonstrations~\cite{sutton2018reinforcement}. However, applying RL to GUI-based environments remains fundamentally constrained by the absence of reliable reward signals~\cite{sumyk2025areyetvisionbasedjudge}. Unlike robotics simulators~\cite{tang2024deepreinforcementlearningrobotics} or game environments~\cite{koyamada2024pgxhardwareacceleratedparallelgame}, where success criteria are explicitly defined and machine-readable, desktop applications rarely expose clear indicators of task completion. Instead, task objectives are typically implicit, visually grounded, and highly unstructured~\cite{xie2024osworldbenchmarkingmultimodalagents}.

As a result, prior work often relies on brittle, task-specific heuristics, such as DOM element checks in web-based environments~\cite{qi2025webrltrainingllmweb}, or on manually annotated success labels~\cite{xie2024osworldbenchmarkingmultimodalagents}. These approaches fail to scale across different applications, operating systems, and task distributions, and they substantially limit generalization beyond narrowly defined settings. An alternative strategy is to define task success internally within the computer-use agent itself~\cite{wang2025uitars2technicalreportadvancing}. However, this introduces additional reliability concerns, as the agent's own perception and reasoning are inherently imperfect and are precisely the components that RL aims to improve. Consequently, the lack of a scalable and reliable reward mechanism remains a central bottleneck for effective RL fine-tuning of computer-use agents.

A promising alternative is to employ autonomous evaluators that assess whether a task has been successfully completed. Recent works~\cite{sumyk2025areyetvisionbasedjudge,lin2025cuarewardbenchbenchmarkevaluatingreward,sumyk2026cuaauditmetaevaluationvisionlanguagemodels,rosset2026artbuildingverifierscomputer} demonstrate that evaluators based on Vision-Language Models (VLMs) can achieve high accuracy and generalize across diverse applications. However, these evaluators inevitably introduce noise: they may incorrectly label unsuccessful executions as completed (false positives) or fail to detect success (false negatives). Naively using evaluator outputs as reward signals can therefore introduce bias and lead to unstable reinforcement learning~\cite{wang2018reinforcement,li2024canderecoach}.

In this paper, we address this challenge by introducing a principled framework for integrating autonomous evaluators into reinforcement learning for CUAs. Rather than treating evaluator outputs as ground truth, we explicitly model evaluator noise and derive a statistically grounded, asymptotically unbiased reward estimator. This formulation enables effective RL fine-tuning under imperfect feedback, allowing CUAs to learn from large-scale interaction data without manual labeling.

This work investigates the following research questions:

\begin{itemize}
    \item \textbf{RQ1:} Can autonomous VLM-based evaluation serve as a scalable reward signal for RL fine-tuning of Computer-Use Agents?
    \item \textbf{RQ2:} Does modeling and correcting evaluator noise improve policy learning compared to using raw evaluator rewards?
\end{itemize}

To address these questions, we propose a principled framework that integrates autonomous evaluators into the reinforcement-learning loop by explicitly modeling their error characteristics. Our approach derives a statistically grounded, asymptotically unbiased reward estimator that corrects evaluator-induced noise and enables stable policy optimization. Empirically, we demonstrate that this framework yields improvements in robustness and task success across three operating systems, Windows, macOS, and Linux, without relying on manual annotations or task-specific heuristics. The implementation is available at \url{https://github.com/martasumyk/rl_with_autonomous_feedback}.

\section{Related Work}

\subsection{Computer-Use Agents}

Computer-Use Agents (CUAs) are end-to-end autonomous systems that complete natural-language tasks by perceiving the rendered desktop GUI, typically via screenshots, and executing actions such as clicking, typing, scrolling, and dragging~\cite{sager2025comprehensivesurveyagentscomputer}. Recent CUA architectures combine vision-language reasoning with explicit action grounding to plan and carry out long-horizon workflows across diverse applications and operating systems~\cite{liu2025infiguiagentmultimodalgeneralistgui,sun2025seagentselfevolvingcomputeruse,wang2025uitars2technicalreportadvancing,qin2025uitarspioneeringautomatedgui}. 

Unlike API-based or function-calling agents that require explicit service integrations, CUAs adopt a service-agnostic interaction model: they perceive and manipulate software exclusively through its rendered interface. This design enables interaction with arbitrary applications without bespoke engineering and allows a single agent to generalize across different software applications and operating systems~\cite{sun2025seagentselfevolvingcomputeruse,sager2025comprehensivesurveyagentscomputer}. Consequently, CUAs are increasingly regarded as a promising foundation for general-purpose computer automation.

However, this generality introduces significant challenges for reasoning and verification. Because CUAs rely solely on visual feedback, they are susceptible to silent or partial failures caused by unexpected interface states, asynchronous rendering, visual occlusions, or subtle distribution shifts in UI layouts~\cite{gur2023browsergym,humphreys2024webarena,li2024seeact}. Furthermore, many real-world tasks lack explicit, machine-readable success criteria, making it difficult to reliably determine whether an agent has truly completed the intended objective~\cite{sumyk2025areyetvisionbasedjudge}. 

Moreover, state-of-the-art desktop agents achieve success rates of only around $60\%$ on OSWorld~\cite{xie2024osworldbenchmarkingmultimodalagents} and approximately $40\%$ on macOSWorld~\cite{yang2025macosworldmultilingualinteractivebenchmark}. This performance gap highlights the need for more robust training and fine-tuning methods for CUAs.

\subsection{Training and Fine-Tuning of Computer-Use Agents}

Training CUAs typically combines supervised learning with reinforcement learning~\cite{lai2025computerrlscalingendtoendonline,wang2025uitars2technicalreportadvancing}. Many systems first use behavioral cloning to map visual observations and instructions to low-level GUI actions from human demonstrations or scripted trajectories~\cite{gur2023browsergym,li2024seeact,humphreys2024webarena}. This approach is effective for short-horizon action prediction and instruction following, but it often struggles with compounding errors and long-horizon tasks that require recovery, exploration, and verification~\cite{humphreys2024webarena}.

To address these limitations, recent work explores RL-based fine-tuning to improve robustness and task success beyond demonstrations~\cite{liu2025infiguiagentmultimodalgeneralistgui,sun2025seagentselfevolvingcomputeruse,qin2025uitarspioneeringautomatedgui}. In practice, RL pipelines either operate in environments where rewards can be programmatically defined, such as synthetic web interfaces~\cite{gur2023browsergym,humphreys2024webarena}, or rely on task- and platform-specific heuristics, such as DOM parsing or string matching, that do not transfer to general desktop settings~\cite{xie2024osworldbenchmarkingmultimodalagents}. More broadly, existing approaches remain limited by the dependence on a reliable task-completion signal, which is rarely available in real-world desktop GUIs and motivates research on autonomous evaluation.

\subsection{Autonomous Evaluation}

Autonomous evaluation seeks to determine whether an agent has successfully completed a user's instruction based solely on the observed GUI state, producing feedback suitable for both benchmarking and learning~\cite{pan2024autonomousevaluationrefinementdigital}. In real-world desktop environments, task success is rarely accompanied by explicit, machine-readable signals, as objectives are often implicit and visually grounded~\cite{sumyk2025areyetvisionbasedjudge}. Consequently, many existing benchmarks and training pipelines rely on human verification of final states, which is costly, time-consuming, and difficult to scale across diverse applications and operating systems~\cite{xie2024osworldbenchmarkingmultimodalagents}. 

Recent work proposes autonomous evaluators, typically based on Vision-Language Models (VLMs), that assess task completion by jointly analyzing the final GUI state and the natural-language instruction, and output a binary success judgment~\cite{sumyk2025areyetvisionbasedjudge,lin2025cuarewardbenchbenchmarkevaluatingreward}. These evaluators enable scalable, automated success labeling and can serve as reward or feedback providers for agent improvement. However, their predictions are imperfect and can include false positives and false negatives. Therefore, naively treating evaluator outputs as ground-truth rewards can bias learning and destabilize policy optimization~\cite{wang2018reinforcement}. This motivates methods that explicitly model evaluator error and incorporate noise-aware reward correction, which is the focus of our approach.

\subsection{Reinforcement Learning with Noisy Feedback}

A broad line of research studies reinforcement learning when the agent does not observe the true reward, but instead receives a noisy proxy produced by measurement artifacts, imperfect annotators, or automated verification systems~\cite{wang2018reinforcement,cai2025reinforcementlearningverifiablenoisy,wang2020reinforcementlearningperturbedrewards}.

Wang et al.~\shortcite{wang2018reinforcement} formalize reward corruption using a confusion matrix over discrete reward values. They show that, given an estimate of the corruption process, one can construct an unbiased reward estimator. This allows standard RL algorithms to recover optimal policies despite observing only corrupted feedback.

Complementary work investigates learning from noisy evaluative signals provided by humans or teachers. For example, CANDERE-COACH~\cite{li2024canderecoach} considers unreliable binary approve/disapprove feedback and proposes online denoising mechanisms that filter feedback before policy updates, demonstrating robustness under substantial noise.

Most closely related to our setting, Cai et al.~\shortcite{cai2025verifiablerewards} study reinforcement learning with verifiable rewards, where policies are trained from noisy binary signals generated by automated verifiers. They model asymmetric false-positive and false-negative errors and derive correction strategies that debias policy-gradient updates by appropriately transforming the observed feedback.

Our method instantiates these ideas for computer-use agents by treating a vision-based task-completion judge as a noisy binary reward channel. Following prior work~\cite{wang2018reinforcement,cai2025verifiablerewards}, we derive a simple correction that yields an asymptotically unbiased reward estimator under a mild separability condition and integrate it directly into PPO. Unlike approaches that denoise feedback at the data level~\cite{li2024canderecoach}, our correction operates at the reward level, making it straightforward to plug into standard policy-gradient fine-tuning.

\section{Methodology}
\label{sec:method}

We formulate a CUA as a Markov Decision Process $\mathcal{M} = (\mathcal{S}, \mathcal{A}, P, r, \gamma)$~\cite{mdp}. We consider a collection of $m$ tasks indexed by $i \in \{1,\dots,m\}$, where each task defines an episode with a fixed natural-language instruction $d_i$ and horizon $T$.

At timestep $t$ of task $i$, the agent observes the current GUI and executes a single atomic interaction. The state is defined as $s_{i,t} = (x_{i,t}, d_i)$, where $x_{i,t}$ denotes the rendered screen image and $d_i$ the task description, fixed throughout the episode. The action space $\mathcal{A}$ consists of low-level GUI operations: \texttt{click}, \texttt{type}, \texttt{scroll}, and \texttt{drag}. A trajectory for task $i$ is:
\begin{equation}
    \tau_i = (s_{i,0}, a_{i,0}, \dots, s_{i,T}).
\end{equation}

\subsection{Dataset}

We construct a synthetic dataset of computer-use tasks spanning $42$ applications for each OS, including built-in macOS applications and functionally similar applications on Linux and Windows. For each application, we define $60$ natural-language task descriptions, resulting in a total of $7{,}560$ tasks that cover a broad range of GUI interaction patterns across operating systems.

For each application, tasks are randomly partitioned into three disjoint splits of $20$ tasks each. Two splits, a total of $40$ tasks per application and two-thirds of the dataset, are used for reinforcement-learning fine-tuning, where agent rollouts are collected and PPO updates are performed using evaluator-based rewards. The remaining split, one-third of the dataset, is reserved for evaluator calibration and is used to estimate the evaluator's true-positive, false-positive, true-negative, and false-negative rates. This separation ensures that evaluator calibration and policy optimization are conducted on disjoint task sets, preventing information leakage.

In addition to our dataset, we incorporate task descriptions from three existing GUI interaction datasets, OmniAct~\cite{kapoor2024omniactdatasetbenchmarkenabling}, GUI-World~\cite{chen2025guiworldvideobenchmarkdataset}, and GUIDE~\cite{chawla2024guidegraphicaluserinterface}, to further diversify the task distribution used for RL fine-tuning. From these datasets, we use only the natural-language task specifications as input to the agent. A summary of all datasets and their task characteristics is provided in Table~\ref{tab:dataset-summary}.

In addition to task definitions, the dataset includes execution logs of a base computer-use agent, UI-TARS~\cite{qin2025uitarspioneeringautomatedgui}, and of agents fine-tuned with reinforcement learning. Each log records a complete agent trajectory, including screenshots, executed actions, and intermediate reasoning steps. For evaluation purposes, task outcomes in the evaluator calibration and RL test splits are annotated with ground-truth success labels, which are used exclusively for evaluator assessment and final benchmarking and are never exposed to the agent during training.

\begin{table*}[t]
\centering
\begin{tabular}{lcc}
\toprule
\textbf{Dataset} & \textbf{Platforms} & \textbf{Task Types} \\
\midrule
OmniAct~\cite{kapoor2024omniactdatasetbenchmarkenabling}
& macOS / Windows / Linux
& Browser, file managers, shell, system utilities \\

GUI-World~\cite{chen2025guiworldvideobenchmarkdataset}
& Web
& Search, email, documents, SaaS dashboards \\

GUIDE~\cite{chawla2024guidegraphicaluserinterface}
& Web
& Enterprise dashboards, productivity tools \\

Ours
& macOS / Windows / Linux
& OS utilities, file I/O, app navigation, settings \\
\bottomrule
\end{tabular}
\caption{
Summary of datasets whose task descriptions are used as input for RL fine-tuning.
All datasets provide natural-language task specifications for GUI interaction; our dataset additionally includes native desktop tasks explicitly designed for autonomous evaluation and reward correction.
}
\label{tab:dataset-summary}
\end{table*}

\subsection{Autonomous Evaluation}

For autonomous evaluation, we use the Qwen2-VL-7B model~\cite{wang2024qwen2vlenhancingvisionlanguagemodels}, motivated by its strong multimodal reasoning and empirically demonstrated superiority as a vision-based evaluator among open-source models~\cite{sumyk2025areyetvisionbasedjudge,lin2025cuarewardbenchbenchmarkevaluatingreward}. Given a task description $d_i$ and the final GUI state $x_{i,T}$, the evaluator outputs a binary signal:
\begin{equation}
    \tilde r_i \in \{0,1\},
\end{equation}
where $\tilde r_i = 1$ indicates that the task is judged completed. The evaluator operates in a zero-shot setting and is fully decoupled from the acting agent: it observes neither the agent's action history nor its internal reasoning, basing its judgment solely on the final screenshot and instruction.

\subsection{Reinforcement Learning Fine-Tuning}

\begin{figure*}[t]
    \centering
    \includegraphics[width=0.82\textwidth]{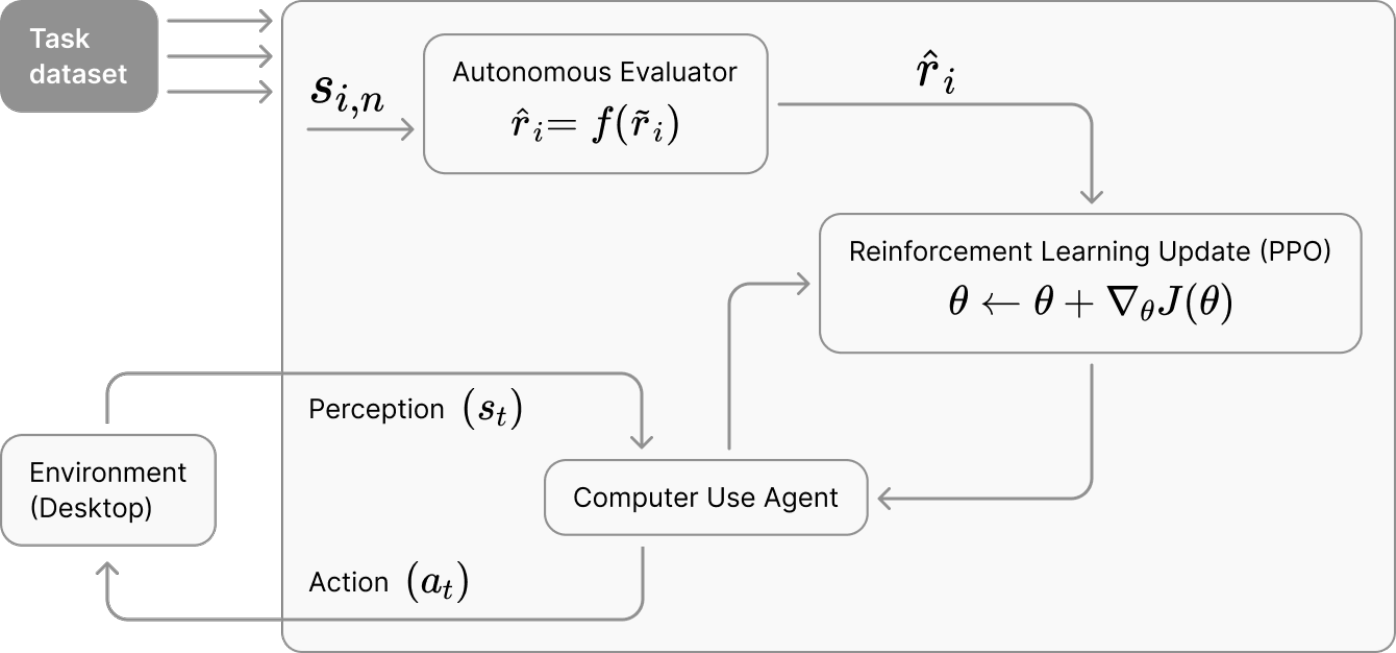}
    \caption{
    Overview of our RL fine-tuning pipeline with autonomous, noise-corrected rewards.
    For each task instance, the computer-use agent interacts with the desktop environment, observing the GUI state $s_t$ and executing actions $a_t$.
    A vision-language evaluator predicts a noisy binary completion signal $\tilde r$ from the final observed state, a screenshot and instruction, which we transform into a corrected reward $\hat r = f(\tilde r)$ using estimated false-positive and false-negative rates.
    PPO then updates the agent parameters $\theta$ using $\hat r$, closing the interaction--evaluation--learning loop.
    }
    \label{fig:methodology-schema}
\end{figure*}

We now describe how we fine-tune a pre-trained CUA using reinforcement learning with world feedback from the autonomous evaluator. Figure~\ref{fig:methodology-schema} provides an overview of the full pipeline.

\paragraph{Terminal Reward and Noise Model.}
In our setting, reward is terminal-only: the true task-completion signal $r_i^\star \in \{0,1\}$ is defined only at the end of the episode and is unobserved. Instead, the evaluator provides a noisy binary judgment $\tilde r_i$ based on the final GUI state $x_{i,T}$.

We characterize evaluator noise using conditional error rates:
\begin{equation}
    e_+ := \Pr(\tilde r_i = 0 \mid r_i^\star = 1) \quad\text{(false negative)},
\end{equation}
\begin{equation}
    e_- := \Pr(\tilde r_i = 1 \mid r_i^\star = 0) \quad\text{(false positive)}.
\end{equation}

We define the separability constant:
\begin{equation}
    D := 1 - e_+ - e_-,
\end{equation}
and assume $D > 0$, ensuring that the evaluator provides informative feedback beyond random guessing.

\paragraph{Noise-Corrected Reward Estimator.}
Since $\tilde r_i$ is binary, its conditional expectation satisfies:
\begin{equation}
    \mathbb{E}[\tilde r_i \mid r_i^\star] = e_- + D \cdot r_i^\star.
\end{equation}
Solving for $r_i^\star$ yields the corrected reward estimator:
\begin{equation}
    \hat r_i := \frac{\tilde r_i - e_-}{D},
\end{equation}
which satisfies:
\begin{equation}
    \mathbb{E}[\hat r_i \mid \tau_i] = r_i^\star,
\end{equation}
and is therefore an asymptotically unbiased estimator of true task completion.

In practice, error rates $e_+$ and $e_-$ are unknown and must be estimated from the held-out calibration split. Let $\hat e_+$ and $\hat e_-$ denote empirical estimates and $\hat D := 1 - \hat e_+ - \hat e_-$. The resulting plug-in estimator is:
\begin{equation}
    \hat r_i^{\,\sharp} := \frac{\tilde r_i - \hat e_-}{\hat D},
\end{equation}
which is asymptotically unbiased as calibration data grows and $\hat D$ remains bounded away from zero.

\paragraph{Policy Optimization.}
Since reward is terminal-only, the discounted return for task $i$ is:
\begin{equation}
    \hat R_i := \gamma^{T}\, \hat r_i,
\end{equation}
and the RL objective is:
\begin{equation}
    J(\theta) := \mathbb{E}_{i \sim \mathcal{D},\, \tau_i \sim \pi_\theta}\!\left[\hat R_i\right].
\end{equation}

We optimize $J(\theta)$ using Proximal Policy Optimization (PPO)~\cite{schulman2017proximalpolicyoptimizationalgorithms}. The actor parameterizes a stochastic policy $\pi_\theta(a \mid s)$ over GUI actions; the critic estimates the value function $V_\phi(s)$. Advantage estimates are computed by propagating the terminal corrected reward $\hat r_i$ backward through the trajectory, enabling stable policy-gradient updates despite evaluator-induced noise. By explicitly correcting evaluator errors, our approach provides a reliable world feedback signal for RL without requiring ground-truth labels or task-specific heuristics.

\section{Results and Evaluation}
\label{sec:results}

We evaluate our approach along three dimensions:
(i) the accuracy and error profile of the autonomous evaluator used to generate reward signals;
(ii) the effect of evaluator-based rewards on PPO fine-tuning, comparing raw evaluator feedback with the proposed noise-corrected reward estimator; and
(iii) an ablation comparing per-OS fine-tuning against a single unified model trained across operating systems.

\subsection{Autonomous Evaluation}
\label{sec:results-eval}

We first quantify the reliability of the vision-based evaluator, since its error rates directly determine the reward correction in Section~\ref{sec:method}. Table~\ref{tab:confusion-all} reports normalized confusion matrices on the held-out evaluator split for macOS, Windows, and Linux. While the evaluator achieves high accuracy overall, its behavior varies by OS.

These asymmetries are important for reinforcement learning: false positives are particularly harmful because they can assign high reward to unsuccessful trajectories, reinforcing incorrect behaviors; false negatives, in contrast, primarily reduce the effective reward frequency and can slow learning. We therefore estimated $(e_+, e_-)$ per OS from the evaluator split and used these values to compute the corrected reward estimator $\hat r_t$ during PPO fine-tuning.

\begin{table*}[t]
\centering
\begin{tabular}{cc|cc|cc|cc}
\toprule
& & \multicolumn{6}{c}{\textbf{Predicted}} \\
& & \multicolumn{2}{c|}{\textbf{macOS}} 
& \multicolumn{2}{c|}{\textbf{Windows}} 
& \multicolumn{2}{c}{\textbf{Linux}} \\
\cmidrule(lr){3-4} \cmidrule(lr){5-6} \cmidrule(lr){7-8}
& & Pos & Neg & Pos & Neg & Pos & Neg \\
\midrule
\parbox[t]{2mm}{\multirow{2}{*}{\rotatebox[origin=c]{90}{\textbf{True}}}}
& Pos & \cellcolor{green!30}\textbf{0.5024} & \cellcolor{red!20}0.1060 
      & \cellcolor{green!50}\textbf{0.6929} & \cellcolor{red!20}0.1226 
      & \cellcolor{green!15}\textbf{0.2488} & \cellcolor{red!30}0.2690 \\
& Neg & \cellcolor{red!20}0.0488 & \cellcolor{green!30}\textbf{0.3429} 
      & \cellcolor{red!20}0.0583 & \cellcolor{green!15}\textbf{0.2488} 
      & \cellcolor{red!20}0.1226 & \cellcolor{green!30}\textbf{0.3595} \\
\bottomrule
\end{tabular}
\caption{
Normalized confusion matrices of the autonomous evaluator across operating systems.
Green cells indicate correct predictions, red cells indicate misclassifications.
}
\label{tab:confusion-all}
\end{table*}

\subsection{Reinforcement Learning Fine-Tuning}

We evaluate reinforcement learning fine-tuning across three desktop environments: macOS, Windows, and Linux, using macOSWorld~\cite{yang2025macosworldmultilingualinteractivebenchmark}, Windows Agent Arena~\cite{bonatti2024windowsagentarenaevaluating}, and OSWorld~\cite{xie2024osworldbenchmarkingmultimodalagents}, respectively.
Results are summarized in Table~\ref{tab:main-results}.

We compare five training configurations:
(1) a zero-shot baseline,
(2) PPO fine-tuning with raw binary evaluator rewards $\tilde r_t$ using a unified cross-OS model,
(3) PPO fine-tuning with the corrected reward estimator $\hat r_t$ using a unified model,
(4) PPO fine-tuning with raw evaluator rewards using per-OS models, and
(5) PPO fine-tuning with the corrected reward estimator using per-OS models.

\begin{table*}[t]
\centering
\begin{tabular}{lccc}
\toprule
\textbf{Training Setup}
& \textbf{macOSWorld} 
& \textbf{WindowsAgentArena} 
& \textbf{OSWorld} \\
\midrule
Zero-shot baseline
& 0.084 & 0.331 & 0.283 \\

RL + raw evaluator reward ($\tilde r_t$), unified model
& 0.129 & 0.383 & 0.385 \\

RL + corrected reward estimator ($\hat r_t$), unified model
& 0.144 & \underline{0.403} & 0.399 \\

RL + raw evaluator reward ($\tilde r_t$), per-OS models
& \underline{0.149} & 0.357 & \underline{0.418} \\

RL + corrected reward estimator ($\hat r_t$), per-OS models
& \textbf{0.203} & \textbf{0.442} & \textbf{0.432} \\
\bottomrule
\end{tabular}
\caption{
Task success rates across desktop benchmarks.
We compare the zero-shot baseline with PPO fine-tuning using
(i) raw binary evaluator rewards and
(ii) the proposed noise-corrected reward estimator,
under both per-OS and unified cross-OS training.
Best results are shown in \textbf{bold}, second-best are underlined.
}
\label{tab:main-results}
\end{table*}

Across all benchmarks, reinforcement learning with evaluator feedback improves performance over the zero-shot baseline.
However, the choice of reward signal and model parameterization substantially affects both final performance and generalization.
Using raw evaluator rewards yields moderate gains in most settings, but these gains are consistently smaller than those obtained with the corrected reward estimator.

The proposed noise-corrected reward $\hat r_t$ achieves the strongest performance across all three benchmarks when combined with per-OS fine-tuning, improving success rates from $0.084 \rightarrow 0.203$ on macOSWorld, $0.331 \rightarrow 0.442$ on Windows Agent Arena, and $0.283 \rightarrow 0.432$ on OSWorld.
These improvements demonstrate that explicitly accounting for evaluator false positives and false negatives yields a substantially more reliable learning signal for PPO.

In contrast, fine-tuning with raw evaluator rewards exhibits inconsistent behavior.
While raw rewards can improve in-domain performance, they underperform the corrected estimator in all settings and, in some cases, reduce generalization.
This effect is most evident on OSWorld, where the corrected reward consistently outperforms raw rewards under both unified and per-OS training.
These results indicate that ignoring evaluator error can introduce bias that negatively impacts reinforcement learning, particularly under distribution shift.

\section{Discussion and Limitations}
\label{sec:limitations}

Firstly, our approach assumes that the evaluator's false-negative and false-positive rates $(e_+, e_-)$ are approximately fixed for a given model and evaluation protocol and can be reliably estimated on a held-out calibration split. In practice, evaluator behavior may be non-stationary: error rates can vary with task type, UI complexity, language phrasing, visual themes, and operating-system-specific interface conventions. Moreover, as the agent improves during training, the distribution of visited GUI states may shift, potentially changing the evaluator's error profile and introducing residual bias when using plug-in estimates $(\hat e_+, \hat e_-)$. A natural extension is to estimate error rates conditionally, for example, per task family or application type.

Another limitation is that our formulation relies on a binary success signal, which is inherently sparse and often effectively terminal. While PPO can learn under sparse rewards, exploration remains challenging for long-horizon GUI tasks with delayed success. The proposed correction improves reward accuracy but does not increase reward density, and therefore cannot by itself resolve exploration difficulties. Incorporating shaping signals, such as intermediate evaluator judgments, progress estimators, or subgoal completion signals, may substantially improve sample efficiency, but would require additional calibration and may introduce new sources of bias.

The effectiveness of the correction also depends on the quality of the calibration set used to estimate $(e_+, e_-)$. With limited calibration data, statistical uncertainty in these estimates can propagate into the corrected reward and increase the variance of policy-gradient updates, particularly when the separability term $D = 1 - e_+ - e_-$ is small and the correction amplifies noise. In such regimes, conservative strategies such as clipping the corrected reward, shrinking estimates toward priors, or abstaining when evaluator confidence is low may be necessary to maintain training stability.

In addition, the corrected reward $\hat r_t = \frac{(\tilde r_t - \hat e_-)}{\hat D}$ is real-valued and can fall outside the $[0,1]$ range, which alters the scale of advantages and may affect PPO optimization in finite-sample settings. Although policy-gradient methods are invariant to affine reward transformations in expectation, practical stability can still depend on careful normalization and hyperparameter choices.

Finally, vision-based evaluators have inherent blind spots: they may miss semantically correct outcomes that are visually subtle or overestimate success based on superficial visual cues. Binary success labels also ignore partial progress and alternative valid solutions, which are common in real-world GUI tasks. Extending evaluators to produce calibrated confidence scores, multi-level outcomes, or preference-based feedback could provide richer learning signals, but would require revisiting both the noise model and the correction mechanism. While our results demonstrate improved generalization to OSWorld, robustness under broader distribution shifts, such as new application versions, different screen resolutions, or accessibility settings, remains an open direction for future work.

\section{Conclusion}
\label{sec:conclusion}

We presented a principled approach for reinforcement-learning fine-tuning of computer-use agents using reward signals produced by an autonomous vision-based evaluator. Our key idea is to treat the evaluator as a binary reward channel and to correct its false-positive and false-negative errors via a simple reward transformation that is asymptotically unbiased under a mild separability condition. This yields a drop-in replacement for raw evaluator rewards that can be used with standard policy-gradient methods such as PPO.

Empirically, PPO fine-tuning with evaluator feedback improves over the zero-shot UI-TARS baseline across operating systems. While fine-tuning with the raw evaluator signal can help on in-domain tasks, we find that explicitly correcting evaluator noise produces more consistent gains, especially on macOS and Windows, and improves transfer to the OSWorld benchmark, where naive use of raw rewards slightly degrades performance. Finally, our ablation comparing per-OS and unified fine-tuning suggests that OS-specific policies provide more reliable generalization overall, supporting the use of per-OS models in our main experiments.

Overall, these results indicate that autonomous evaluation can serve as a scalable supervision signal for RL in GUI environments, provided that evaluator noise is explicitly modeled and corrected. We view this as a step toward practical RL fine-tuning pipelines for general desktop agents that learn from interaction without requiring manual success labels.

\bibliographystyle{named}
\bibliography{ijcai26}

\end{document}